\documentclass[letterpaper, 10 pt, conference]{support/ieeeconf}
\IEEEoverridecommandlockouts
\usepackage{cite}
\usepackage{amsmath,amssymb,amsfonts}
\usepackage{algorithmic}
\usepackage{graphicx}
\usepackage{textcomp}
\usepackage{hyperref}
\usepackage{booktabs}
\usepackage{multirow}
\usepackage{makecell}
\usepackage{etoolbox}
\usepackage{url}
\usepackage{booktabs}
\usepackage[ruled,linesnumbered]{algorithm2e}
\usepackage{amsmath}
\usepackage{float}

\makeatletter
\makeatletter
\def\endthebibliography{%
	\def\@noitemerr{\@latex@warning{Empty `thebibliography' environment}}%
	\endlist
}

\patchcmd{\@makecaption}
  {\scshape}
  {}
  {}
  {}
\makeatletter
\patchcmd{\@makecaption}
  {\\}
  {.\ }
  {}
  {}
\makeatother
\def\tablename{Table}   

\def\BibTeX{{\rm B\kern-.05em{\sc i\kern-.025em b}\kern-.08em
    T\kern-.1667em\lower.7ex\hbox{E}\kern-.125emX}}
\begin{document}
\title{Topology-Driven Trajectory Optimization for Modelling Controllable Interactions Within Multi-Vehicle Scenario\\
	\thanks{$^1$The Academy for Engineering and Technology, Fudan University, Shanghai, China.(\emph{Corresponding Authors: Zhongxue Gan, Wenchao Ding)}}
	\thanks{$^2$School of Automotive Studies, Tongji University, Shanghai, China.}
	\thanks{This work was supported by Shanghai Municipal Science and Technology Major Project(No.2021SHZDZX0103).}
	\thanks{This work was supported by the Shanghai Engineering Research Center of AI \& Robotics, Fudan University, China, and the Engineering Research Center of AI \& Robotics, Ministry of Education, China.}
    \thanks{E-mail: {\tt\small \{cjma24\}@m.fudan.edu.cn}}
}
\author{Changjia Ma$^1$, Yi Zhao$^1$, Zhongxue Gan$^1$, Bingzhao Gao$^2$ and Wenchao Ding$^1$}

\maketitle

\begin{abstract}
Trajectory optimization in multi-vehicle scenarios faces challenges due to its non-linear, non-convex properties and sensitivity to initial values, making interactions between vehicles difficult to control. In this paper, inspired by topological planning, we propose a differentiable local homotopy invariant metric to model the interactions. By incorporating this topological metric as a constraint into multi-vehicle trajectory optimization, our framework is capable of generating multiple interactive trajectories from the same initial values, achieving controllable interactions as well as supporting user-designed interaction patterns. Extensive experiments demonstrate its superior optimality and efficiency over existing methods. We will release open-source code to advance relative research\footnote{https://github.com/Fudan-MAGIC-Lab/Interactive-Traj-Opt}.
\end{abstract}

\section{Introduction}
Nowadays, multi-vehicle scenarios are increasingly common in daily life, such as autonomous driving vehicles at the intersections interacting with other vehicles and pedestrians.
Multi-vehicle systems are also utilized in various applications, ranging from logistics and delivery services to traffic management and surveillance, significantly improving productivity.
All these applications depend on reliable trajectory planning algorithms, among which the most commonly adopted are the optimization-based methods, to generate smooth and safe trajectories.
However, trajectory optimization in multi-vehicle scenarios faces significant challenges:
1) High-dimensional, non-convex and non-linear: The final solution heavily depends on initial values, making it difficult to fall into reasonable local minima.
2) Uncontrollable interactions: Existing methods lack the ability to control interactions between vehicles during optimization, limiting their practical applicability.
Inspired by topological planning for robots, we propose a homotopy invariant metric to model the interactions within multi-vehicle scenarios.
By incorporating this metric as a constraint into the optimization framework, we enable generating trajectories under multiple interactive patterns from the same initial values, achieving controllable interactions. 

In multi-vehicle scenarios like autonomous driving, vehicles face complex interactions with other vehicles and obstacles. 
Human drivers anticipate trajectories to assess possible interactions and make decisions, such as overtaking, yielding, following, or navigating around obstacles. 
However, translating these behaviors into algorithmic frameworks is challenging, as it requires precise mathematical modeling. 
While topology-based methods have advanced the analysis of spatial-temporal interactions, their metrics are often discrete and non-differentiable, limiting controllability over interactions during planning. 
To address this, we propose a differentiable local homotopy invariant metric that continuously adapts to trajectory topology changes, enabling better control of vehicle interactions.

\begin{figure}[t]
    \vspace{0.2cm}
    \centering
    \setlength{\abovecaptionskip}{-0.00cm}
    \includegraphics[width=8.5cm]{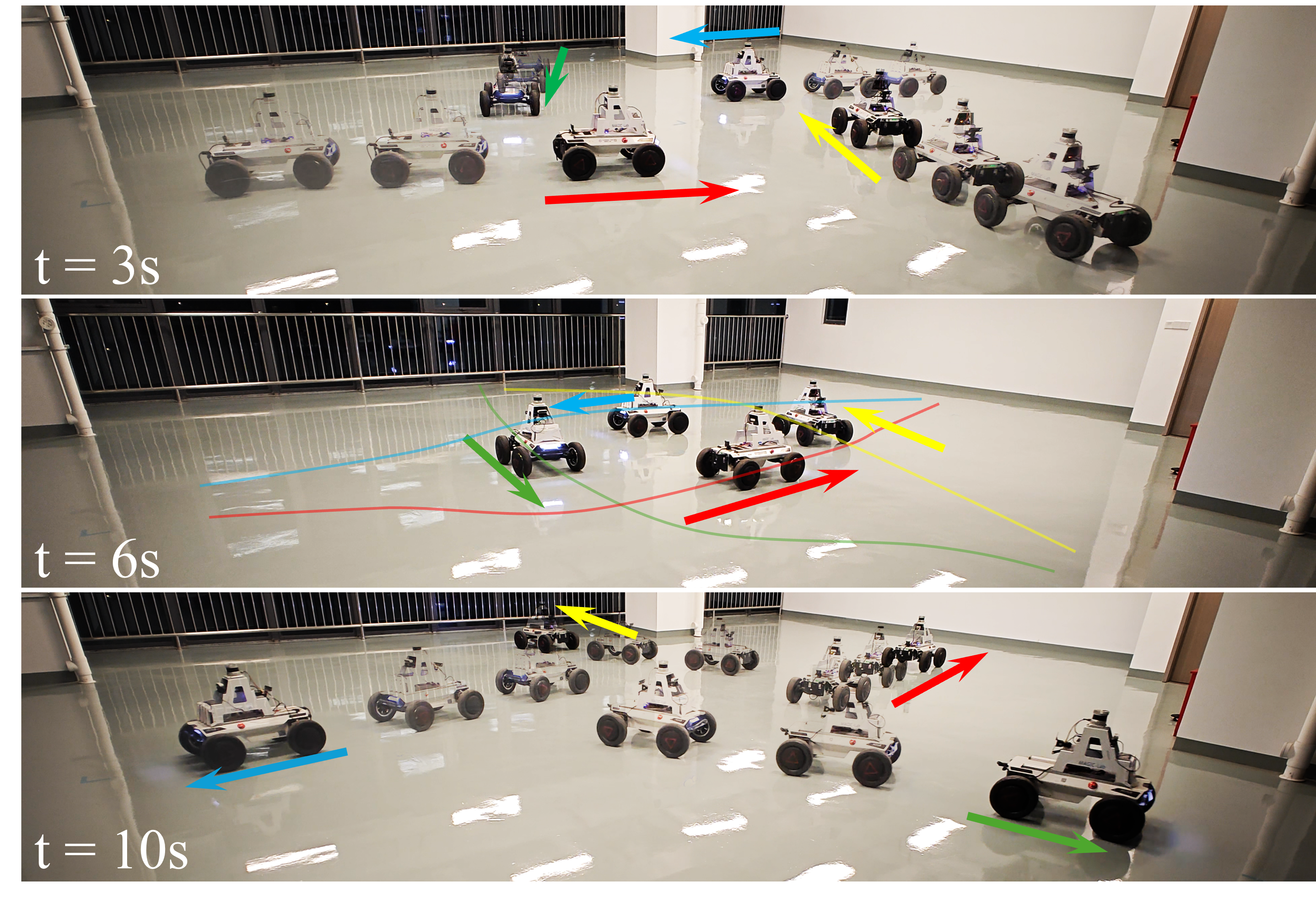}
    \caption{Snapshots of four vehicles navigating at the interaction area. The arrows point to the directions that the vehicles are moveing towards. The colored curves are trajectories of each vehicle. In order to avoid collision, the vehicles move counterclockwise relative to each other at t=6s. This illustrates an interaction pattern within multi-vehicle scenario.}
    \label{fig:real world experiments}
    \vspace{-0.35cm}
\end{figure}

Trajectory optimization in multi-vehicle scenarios is a challenging, high-dimensional, non-convex problem involving kinodynamic and collision avoidance constraints. 
Solvers often converge to suboptimal local minima, which are highly sensitive to initial values and correspond to different interaction patterns. 
While most existing methods neglect interaction relationships, some attempt to model interactions as topological constraints. 
However, their non-differentiable metrics pose great burdens on the solvers. 
In this work, we address this by introducing the proposed continuous and differentiable topological metric as an inequality constraint, transforming the constrained problem into an unconstrained one. 
This enables optimizing multiple interaction patterns from the same initial value, significantly enhancing controllability. \looseness=-1

\begin{figure}[t]
	\centering
	\includegraphics[width=6.0cm]{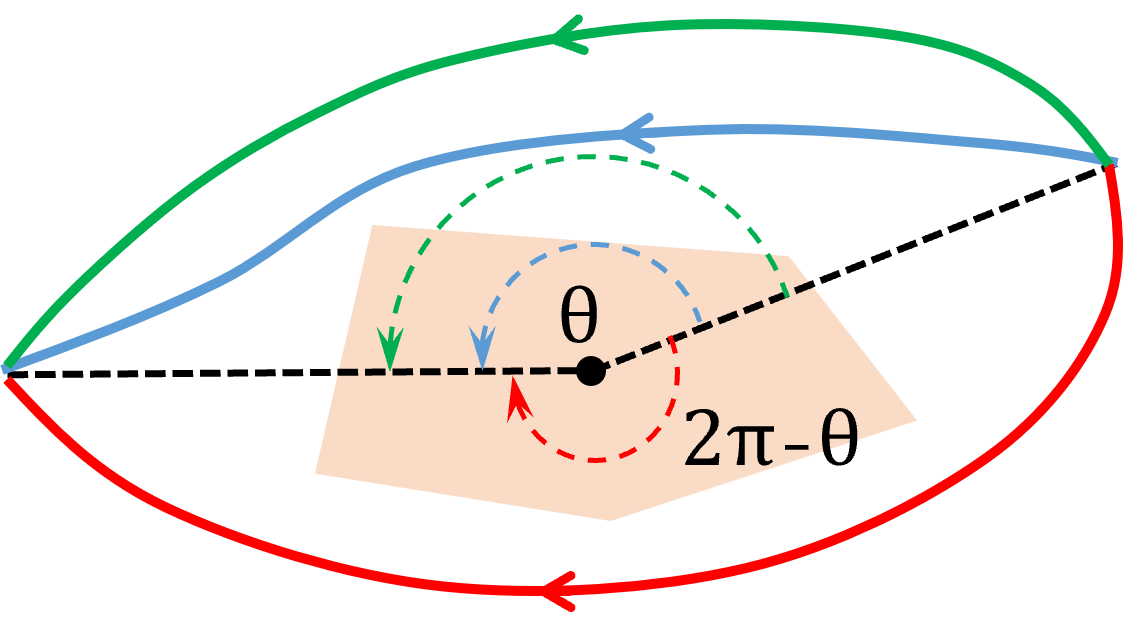}
	\setlength{\abovecaptionskip}{-0.2cm}
	\caption{Illustration of trajectories under different homotopy classes with different winding angles. The green and blue trajectories navigate above the obstacle, with a common winding angle of $\theta$. The red trajectory navigates below the obstacle, with a winding angle of $2\pi-\theta$. The winding angles are represented by the dashed lines with arrows. All trajectories navigate from right to left, sharing the same start and goal.}
	\label{fig:topology nondifferential}
	\vspace{-0.5cm}
\end{figure}

Trajectory optimization with controllable interactions is crucial for motion planning in multi-vehicle scenarios. 
By enabling the generation of multiple interaction patterns from the same initial values, our approach avoids the computational burden of finding multiple initial conditions and allows for user-designed interaction patterns, providing greater flexibility for decision-making. 
Unlike existing methods that rely on discrete and non-differentiable topological metrics, we propose a continuous and differentiable topological metric to model interactions. 
This metric is formulated as an inequality constraint and incorporated into the optimization framework.
Then we transform the constrained problem into an unconstrained one, enabling precise control over vehicle interactions.
Extensive experiments in simulation and real-world scenarios demonstrate that our framework outperforms existing methods in both optimality and computational efficiency. 
Contributions of this paper are summarized as below:
\begin{enumerate}
    \item We propose a simple and differentiable topological metric to model interactions in multi-vehicle scenarios, addressing the limitations of non-differentiable metrics in existing works.
    \item We incorporate the differentiable topological metric into trajectory optimization framework and transform it into an unconstrained one, making it easier to solve and realizing controllable interactions.
    \item We conduct extensive experiments and benchmark comparisons to validate the superior optimality and efficiency of our framework.
    \item We will release our code as open-source packages to serve the research community.
\end{enumerate}

\section{RELATED WORKS}
\subsection{Topology-Driven Trajectory Planning}
The interactions between trajectories and obstacles naturally lend themselves to topological analysis.
Early work by \cite{bhattacharya2010search} introduced homotopy and homology classes for trajectory planning, proposing homotopy invariants, which are essentially topological metrics to describe topological structures.
Subsequent studies expanded on this, with various topological metrics emerging.

For instance, \cite{7182335,sontges2017computing,bhattacharya2018path,6425970} used lines or planes extending from obstacles to divide the planning space into zones, assigning unique labels to trajectories based on their traversal of these zones.
Similarly, braid theory \cite{mavrogiannis2022analyzing,liu2024betop,doi:10.1177/02783649231188740} has been employed to model on-road vehicle interactions, representing trajectories as braids with unique labels.
Visibility deformation (VD) roadmaps \cite{jaillet2008path,zhou2021raptor,zhou2022swarm} have also been used to classify trajectories into homotopy classes.
While these methods are intuitive, their reliance on discrete labels makes the topological metrics non-computable, limiting their applicability.

As for computability, \cite{bhattacharya2010search} introduced the $H$-signature, inspired by electromagnetism laws, where obstacles are treated as electric currents and the metric is computed via line integrals of magnetic field intensity along trajectories.
This concept was further utilized by \cite{bhattacharya2012topological,gu2016automated,baselga2024shine}.
While computable, this approach faces challenges in higher-dimensional spaces due to the complexity of accurate integration. 

Simplified versions of the $H$-signature, such as winding numbers or angles, have been proposed and leveraged \cite{inbook,kretzschmar2016socially,roh2021multimodal,mavrogiannis2021hamiltonian,mavrogiannis2022winding,chen2023interactive} for 2D cases. However, these metrics remain non-differentiable, as small trajectory deformations can cause abrupt changes in winding angles.
As shown in Fig. \ref{fig:topology nondifferential}, the winding angle of trajectories at upper side(blue and green) is $\theta$, whereas the lower side(red) is $2\pi - \theta$.
If the trajectory gradually deforms from the blue one, to the green one, then to the red one, there will be a mutation from $\theta$ to $2\pi - \theta$.
This mutation is the origin of discontinuity and non-differentiability. 

In summary, existing methods lack a continuous and differentiable topological metric to describe vehicle interactions. 
To fill this gap, we propose a local topological metric for 2D multi-vehicle scenarios, enabling precise and computable interaction modeling.

\subsection{Trajectory Optimization in Multi-Vehicle Scenario}
Trajectory optimization in multi-vehicle scenarios has been extensively studied. 
Recent works, such as \cite{ma2023decentralized,10285583,10325486}, formulate the problem as an unconstrained optimization, achieving fast and high-quality solutions.
However, these methods converge to random local minima, making it impossible to control the topological structures or interaction patterns.
\cite{de2024topology} introduced topological constraints by restricting waypoints to one side of hyperplanes.
While effective, this approach severely compresses the solution space.
Other frameworks, such as \cite{Li_2021,huang2024spatiotemporal,10533430}, adopt priority-based planning to implicitly model interactions, but their sequential optimization sacrifices optimality.
Additionally, \cite{mavrogiannis2021hamiltonian,mavrogiannis2022winding,chen2023interactive} incorporate winding angles as topological constraints, which naturally fit the problem formulation but suffer from non-differentiability, imposing significant computational burdens.

Despite these efforts, existing methods struggle to control interactions effectively, especially when targeting interactions differ from the initial values.
As a result, they are better suited to maintaining trajectories within a homotopy class rather than transitioning between classes. 
This limitation has led to the use of trajectory samplers, as in \cite{chen2023interactive,de2024topology}, to generate multiple non-homotopic trajectories for parallel optimization.
In this work, we overcome these challenges by incorporating the proposed differentiable topological metric as a constraint into the optimization framework.
By transforming the topological constraint into a differentiable penalty term, the optimization problem is more tractable, enabling controllable interactions from the same initial values.

\section{Preliminaries} 
\label{sec:Preliminaries}
\subsection{Homotopy Invariant in 2D Space}
Homotopy invariant is a mathematical or logical topological metric to discriminate the trajectories with different homotopy classes.
Here we introduce the homotopic condition:

\textbf{\textit{Definition 1:}} homotopic: trajectories sharing the same start and end points, are homotopic if and only if they have the same homotopy invariant\cite{bhattacharya2010search}.

In order to describe the interactions between vehicles in 2D space, we refer to the winding angle, an intuitive and simplified formulation as homotopy invariant.
Centered at the obstacle, connect this center and a point on the trajectory with a straight line, when the point moves from start to end, the straight line swepts over a sector-like area.
The winding angle refers to the center angle of the sector-like area.
As shown in Fig. \ref{fig:topology nondifferential}, three trajectories share the same start and end points.
The green and blue trajectories are located at the upper side(clockwise) of the obstacle, and they have the same winding angle $\theta$, hence they are considered as homotopic.
As for the red trajectory, it moves along the downside(counterclockwise) of the obstacle, with the winding angle of $2\pi-\theta$, is treated as non-homotopic with the green and blue trajectories.

\begin{figure}[t] 
    \centering
    \includegraphics[width=8.8cm]{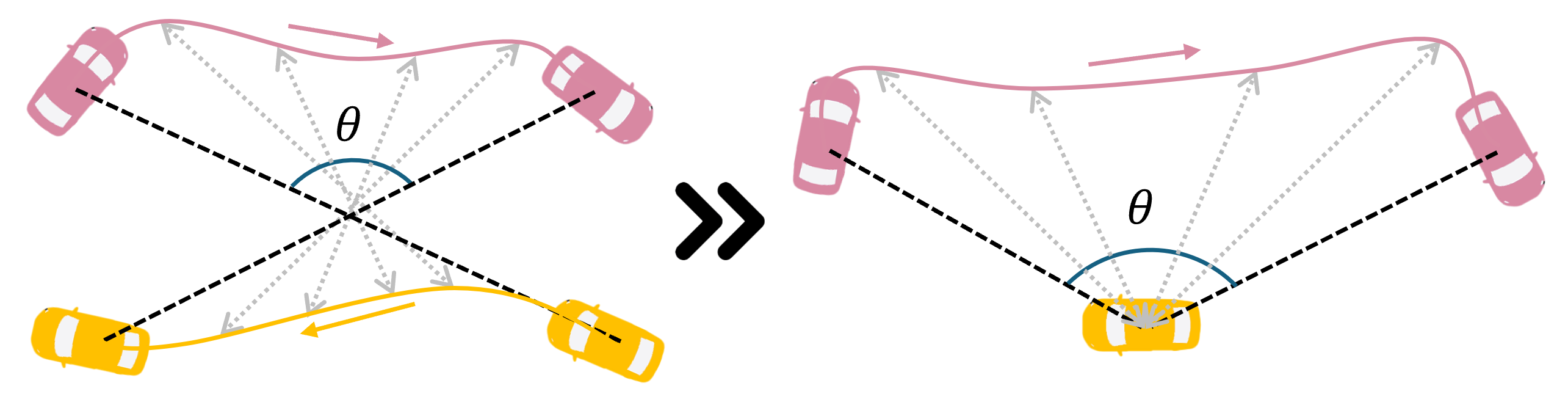}
    \setlength{\abovecaptionskip}{-0.5cm}
    \caption{This figure illustrates that the topological structure between two moving vehicles is equivalent to that between a moving vehicle and a static obstacle. The left figure describes the absolute trajectories of two moving vehicles, whereas the right figure describes the relative trajectory by fixing the yellow vehicle.}
    \label{fig:relative positions}
    \vspace{-0.5cm}
\end{figure}

Noted that the example in Fig. \ref{fig:topology nondifferential} is about trajectories bypassing a static obstacle.
For two moving vehicles, the winding angle can equivalently be acquired by calculating the relative positions of the trajectories, as shown in Fig. \ref{fig:relative positions}.

\subsection{Trajectory Representation and Opimization for Vehicles}
Benefiting from the differential flatness property of the vehicles, all states can be derived from the position $\boldsymbol{p}:=\left[p_{x}, p_{y}\right]^{T}$ as well as its higher order derivatives.
Here we formulate the trajectory as a 2-Demensional 5th order piecewise polynomial, called MINCO\cite{wang2022geometrically}.
Suppose a trajectory consists of $M$ pieces, for the $i$-th piece, the coefficient vector is $\mathbf{c}_i \in \mathbb{R}^{6 \times 1}$, then this piece can be written as:
\begin{align}
\label{equ: trajectory representation}
\begin{gathered}
\boldsymbol{p}_{i}(t) :=\mathbf{c}_{i}^T \boldsymbol{\beta}(t), \\
\boldsymbol{\beta}(t) :=\left[1, t, t^2, t^3, t^4, t^5\right]^T,
\end{gathered}
\end{align}  
where $t\in [0, T_i]$, and $i\in \left\{1,2,...,M \right\}$, $T_i$ is the time duration of the $i$-th piece.

Adopting the trajectory optimization framework from our previous work\cite{ma2023decentralized}, the optimization is transformed into an unconstrained problem by penalty method:
\begin{align}
	\label{equ: optimization formulation}
\min _{\mathbf{c}, \mathbf{T}} &J=\int_0^{T} \boldsymbol{\mu}(t)^T \boldsymbol{\mu}(t) d t+w_T T +S_{\Sigma}(\mathbf{c}, \mathbf{T}),
\end{align}
where $\mathbf{c} = \left[ \mathbf{c}_1,...,\mathbf{c}_M \right] \in \mathbb{R}^{6 \times M}$ is the coefficient matrix, $\mathbf{T}=[T_1, T_2, ...,T_M]\in \mathbb{R}^{M}$ is the time duration vector. $\boldsymbol{\mu}(t)$ denotes the control efforts jerk, and $w_T$ represents the penalty weight on total trajectory duration $T=\sum_{i=1}^{M}T_i$. $S_{\Sigma}(\mathbf{c}, \mathbf{T})$ is the penalty term transformed from inquality constraints such as kinodynamic and collision avoidance constarints.

As shown in Eqa.(\ref{equ: optimization formulation}), by optimizing over the polynomial coefficients $\mathbf{c}$ and time duration $\mathbf{T}$, the goal of the spatial-temporal trajectory optimization problem is to generate a smooth trajectory with the shortest possible time duration, as well as satisfying constraints.
For more details, we refer readers to our previous work\cite{ma2023decentralized}.

\section{Methodology}
In this section, we first introduce a continuous and differentiable local homotopy invariant.
Then we incorporate this metric into the multi-vehicle trajectory optimization problem with a formulation of bi-level optimization framework.
Finally, to avoid the possible local minima stuck between topological and obstacle avoidance constraints, the trajectoroy optimization process is divided into two stages to deal with the constraints respectively.

\subsection{Differentiable Local Homotopy Invariant}
As introduced in Section \ref{sec:Preliminaries}, the winding angle is calculated by sweeping from start to end point of the trajectory.
Apparently, this metric is a global measure, completely modeling the topological relationships.
However, when we talk about the interaction between a vehicle and an obstacle, we are more concerned with the part where the car is very close to the obstacle.
For most part of the trajectory where the vehicle is far from the obstacle, it doesn't matter for the interaction.
Hence, as a homotopy invariant, most part of the winding angle is redundant.

Instead of adopting the winding angle, we use clockwise and counterclockwise directions to describe the interaction.
This strategy is similar to \cite{chen2023interactive} but it still leverages the winding angle.
As analyzed above, we only focus on the trajectory point nearest to the obstacle, and we call it the key point.
The interaction pattern depends on whether the key point moves clockwise or counterclockwise around the obstacle.
Then we introduce how to get this local homotopy invariant in a form of numerical metric.

\begin{figure}[t]
	\centering
	\includegraphics[width=8.7cm]{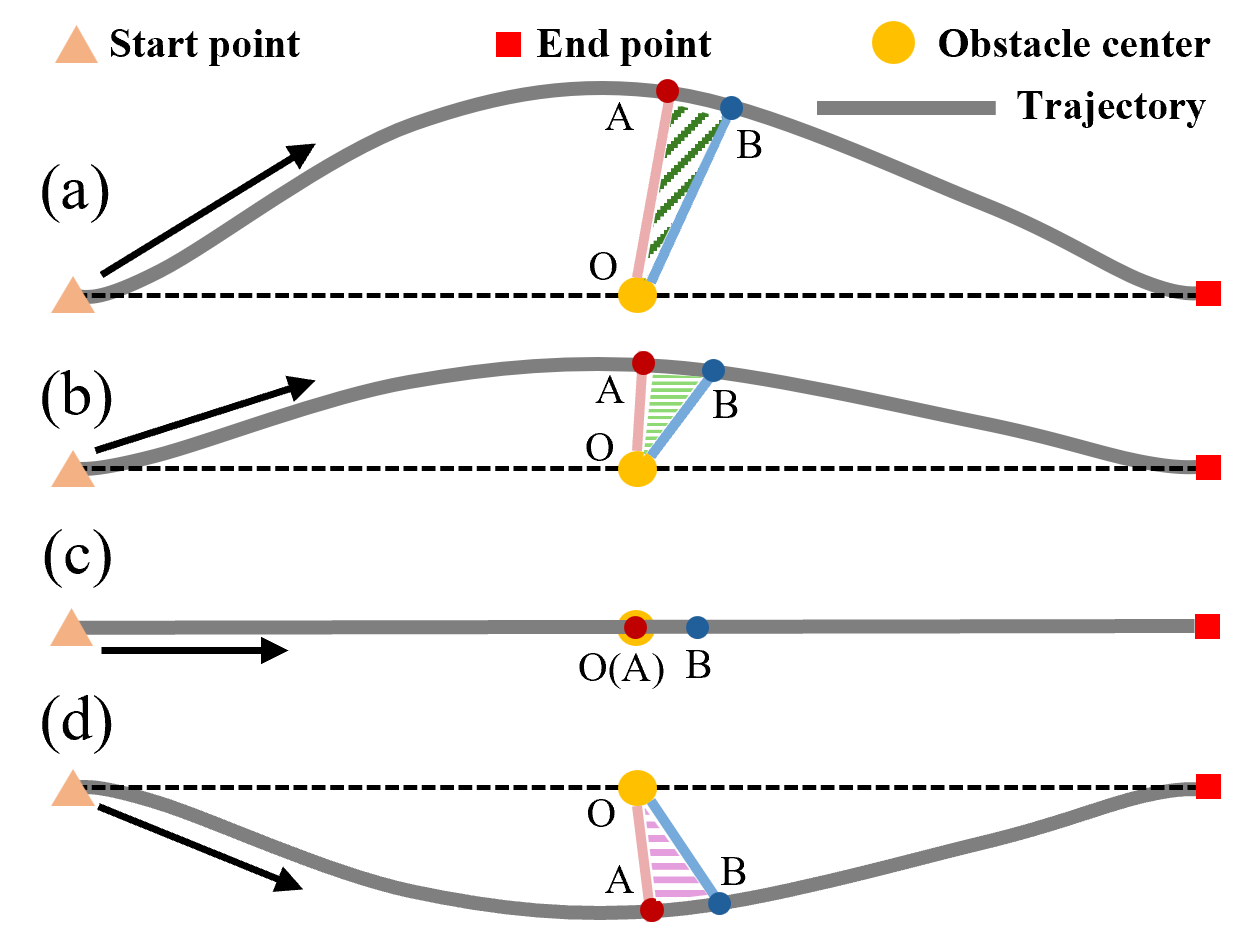}
	\setlength{\abovecaptionskip}{-0.4cm}
	\caption{Illustration of a gradual trajectory deformation process from (a) to (d). For simplicity, the obstacle is represented by its center O. Point A is the key point, and point B is a trajectory point slightly later than point A. In (a) and (b), the key point A moves clockwise around the obstacle to point B. In (c), the key point coincides with the obstacle center. In (d), the key point A moves counterclockwise around the obstacle to point B.}
	\label{fig: topology deformation}
	\vspace{-0.2cm}
\end{figure}

Fig. \ref{fig: topology deformation} illustrates a gradual process of trajectory deformation, where the trajectory gets across the obstacle and changes its topology.
In order to model the topological structure in a continuous manner, we want to find a numerical metric that reflects how far the key point is from the obstacle center.
The closer the key point is to the obstacle, the closer the trajectory is to the boundary of topological transformation.
Fig. \ref{fig: topology deformation}(c) shows the exact boundary of topological change when the key point coincides with the obstacle center.
The first thought to model this distance is the exact distance between obstacle center and key point.
However, this distance is a constant positive value, and it cannot reflect the key point moves clockwise or counterclockwise.
Then we came across the idea inspired by the cross product between vectors.
Suppose the timestamp of key point A is $t^*$, we can find a slightly late trajectory point B at the timestamp of $t^*+\Delta t$.
For simplicity, the obstacle center is the coordinate origin, then we have $\overrightarrow{OA} = [x(t^*), y(t^*)]^T$ and $\overrightarrow{OB} = [x(t^*+\Delta t), y(t^*+\Delta t)]^T$.
Since $\Delta t$ is a minimum value, we can approximate by $\overrightarrow{OB} \approx [x(t^*)+\frac{dx}{dt}\Delta t, y(t^*)+\frac{dy}{dt}\Delta t]^T$, where $\frac{dx}{dt}$ and $\frac{dy}{dt}$ are the trajectory velocity at the key point on x and y direction, respectively.
The physical meaning of the cross product of $\overrightarrow{OA}$ and $\overrightarrow{OB}$ is the area of parallelogram spanned by these two vectors, which is double area of the triangle OAB.
Therefore, we can conduct a cross product:
\begin{align}
2\Delta S = |\overrightarrow{OA} \times \overrightarrow{OB}| = |(x\frac{dy}{dx} - y\frac{dx}{dt})\Delta t|,
\end{align}
where $\Delta S$ is the area of triangle OAB, as the shaded area shown in Fig. \ref{fig: topology deformation}.
Here we reserve the linear main part as the local homotopy invariant metric $\mathcal{M}$:
\begin{align}
	\label{eqa: local homotopy invariant}
	\mathcal{M}(\boldsymbol{p}, \dot{\boldsymbol{p}}) = x\frac{dy}{dx} - y\frac{dx}{dt} = \dot{\boldsymbol{p}}^T \mathbf{B}\boldsymbol{p},
\end{align}
where matrix $\mathbf{B}:=\left[\begin{array}{cc}
	0 & -1 \\
	1 & 0
\end{array}\right]$, $\boldsymbol{p}=[x(t^*),y(t^*)]^T$.

Metric $\mathcal{M}$ represents the rate of change of the swept area by vector $\overrightarrow{OA}$, which can also be considered as the limit area of triangle swept by $\overrightarrow{OA}$ within a minimum period $\Delta t$.
Intuitively, when the key point approaches the obstacle center, the limit area of the swept triangle will tend to zero.
From Eqa.(\ref{eqa: local homotopy invariant}) we can find $\mathcal{M}$ is continuous and its gradient to $\boldsymbol{p}$, $\dot{\boldsymbol{p}}$ can be easily acquired.
Besides, the positive or negative sign of this metric is also significant.
When $\mathcal{M}$ is positive, the key point moves counterclockwise around the obstacle; when negative, clockwise.
Therefore, the proposed metric not only reflects the topology structure, but the value of it also describes how far the key point is from the boundary of topological change, making it continuous and differentiable.

Due to the fact that the proposed homotopy invariant only focuses on the key point, we define local homotopic:

\textbf{\textit{Definition 2:}} local homotopic: trajectories sharing the same start and end points, are local homotopic if and only if their local homotopy invariants are of the same sign, either both positive or both negative.

Next we will formulate the proposed local homotopy invariant as a constraint and incorporate it into the trajectory optimization problem.

\subsection{Bi-Level Trajectory Optimization }
As discussed in Sec.\ref{sec:Preliminaries}, topological structure between vehicles is equivalent to that between a vehicle and a static obstacle, where the only difference is the position and velocity in Eqa.(\ref{eqa: local homotopy invariant}) are relative position and velocity:
\begin{align}
	\label{equ: multi-vehicle homotopy invariant}
	\mathcal{M}(\boldsymbol{p}, \hat{\boldsymbol{p}}, \dot{\boldsymbol{p}}, \dot{\hat{\boldsymbol{p}}}) = (\dot{\boldsymbol{p}} - \dot{\hat{\boldsymbol{p}}})^T \mathbf{B}(\boldsymbol{p} - \hat{\boldsymbol{p}}),
\end{align}
where $\boldsymbol{p}, \hat{\boldsymbol{p}}$ are the positions of ego vehicle and the surround vehicle, respectively.
Since the interaction between vehicles is pairwise, we use ego vehicle and surround vehicle to describe this relationships in the following derivations.

Before trajectory optimization process, we have to determine the interaction patterns between vehicles either by decision-making module or user-design.
The interactions can be expressed by $\eta \in \left\{-1,0,1 \right\}$, respectively means counterclockwise, no interaction, clockwise.
Following \textbf{\textit{Definition 2}}, we aim to optimize the trajectories to be local homotopic with those that satisfy the pre-defined interaction patterns.
That is, if we want to constrain two vehicles to navigate clockwise then $\mathcal{M} \geq 0$ should be guaranteed, and vice versa.
Hence, the topological constraint can be expressed as an inequality constraint:
\begin{align}
	\eta \cdot \mathcal{M} \leq 0.
\end{align}
Then we formulate this constraint into a penalty function:
\begin{align}
\label{equ: topological penalty term}
\mathcal{G}(\boldsymbol{p}, \hat{\boldsymbol{p}}, \dot{\boldsymbol{p}}, \dot{\hat{\boldsymbol{p}}}) = \left\{
\begin{array}{lcl}
\eta \cdot \mathcal{M}(\boldsymbol{p}, \hat{\boldsymbol{p}}, \dot{\boldsymbol{p}}, \dot{\hat{\boldsymbol{p}}}) & {\eta \cdot \mathcal{M} > 0} \\
0  & \eta \cdot \mathcal{M} \leq 0 
\end{array} \right.
\end{align}
In this work, we solve the trajectory optimization problem for all vehicles in a fashion of global planning.
So add this penalty term $\mathcal{G}$ to Eqa.(\ref{equ: optimization formulation}), we get the unconstrained multi-vehicle trajectory optimization problem:
\begin{equation}
\label{equ: multi-vehicle trajectory optimization}
\begin{aligned}
	\min _{\substack{\mathbf{c}^{1},...,\mathbf{c}^{N} \\ \mathbf{T}^{1},...,\mathbf{T}^{N} }} J&=\sum_{i=1}^{N}\int_0^{T^{i}} \boldsymbol{\mu}^i(t)^T \boldsymbol{\mu}^i(t) d t+w_T T^i +S_{\Sigma}(\mathbf{c}^i, \mathbf{T}^i) \\
	&+ \sum_{i=1}^{N}\sum_{j=1\&j\neq i}^{N} w_{t} \cdot \mathcal{G}(\mathbf{c}^{i}, \mathbf{c}^{j}, \mathbf{T}^{i},\mathbf{T}^{j}),
\end{aligned}
\end{equation}
where $N$ is the total number of vehicles, $w_t$ is the penalty weight on topology term.
Letters with superscript $i, j$ represents parameters corresponding to the $i$-th and $j$-th vehicle.

Note that the independent variables of penalty term $\mathcal{G}$ in Eqa.(\ref{equ: topological penalty term}) and Eqa.(\ref{equ: multi-vehicle trajectory optimization}) are different.
This results from the fact that the key points are associated with trajectory parameters.
To be specific, $\boldsymbol{p}$ and $\boldsymbol{\hat{p}}$ are acquired by calculating the minimum distance between two trajectories parameterized with $\mathbf{c},\hat{\mathbf{c}},\mathbf{T},\hat{\mathbf{T}}$. 
For the rest of the mathematical deviations, for simplicity, we only preserve the corresponding parameters with ego vehicle, and deviations with the surround vehicle are symmetrical.
Suppose at the timestamp of $t^*$, ego vehicle is closest to the surround vehicle.
The key point $\boldsymbol{p}^*$ is located at the $k$-th piece of the trajectory, then we have the relative timestamp $\bar{t}$ within this piece:
\begin{align}
	\label{equ: relative time}
	\bar{t} = t^* - \sum_{i=1}^{k-1}T_i.
\end{align}
Following Eqa.(\ref{equ: trajectory representation}), the key point can be represented as:
\begin{align}
	\label{equ: key point calculation}
	\boldsymbol{p}^* = \boldsymbol{p}(\mathbf{c}, \bar{t}(t^*, \mathbf{T})) = \mathbf{c}_k^T \boldsymbol{\beta}(\bar{t}).
\end{align}
Since the key point is the closest point between two vehicles, it is necessary to solve an optimization problem to get $t^*$:
\begin{align}
	\label{equ: inner optimization}
	t^* = \arg\min_t\|\boldsymbol{p}(\mathbf{c},\bar{t}(t,\mathbf{T})) - \hat{\boldsymbol{p}}\|_2^2.
\end{align}
Thus, optimization problem (\ref{equ: inner optimization}) embedded within (\ref{equ: multi-vehicle trajectory optimization}) makes it a bi-level optimization problem, where (\ref{equ: inner optimization}) is the lower-level problem and (\ref{equ: multi-vehicle trajectory optimization}) is the upper-level problem.

To solve the bi-level optimization problem, the critical thing is to acquire the gradients of $t^*$ w.r.t $\mathbf{c}$ and $\mathbf{T}$.
Leveraging the KKT condition to derive the bi-level optimization problem\cite{gould2016differentiating}, for the lower-level problem:
\begin{align}
	\label{equ:lower level problem}
	g(x) = \arg\min_{y}f(x,y) = y^*,
\end{align}
the gradients can be derived as:
\begin{align}
	\label{equ: bi-level gradients}
	\frac{dy^*}{dx} = \frac{dg(x)}{dx} = -\frac{f_{XY}(x,g(x))}{f_{YY}(x,g(x))},
\end{align}
where $f_{XY}$ is the mixed partial derivative of $f$ with respect to $x$ and $y$, and $f_{YY}$ is the partial derivative of $f$ with respect to $y$, taken twice.
This theory perfectly suits the lower-level problem (\ref{equ: inner optimization}), where $\mathbf{c}$ and $\mathbf{T}$, $t$, $\|\boldsymbol{p}(\mathbf{c},\bar{t}(t,\mathbf{T})) - \hat{\boldsymbol{p}}\|_2^2$ correspond to $x,y,f$ in (\ref{equ:lower level problem}), respectively.

Let $f=\|\boldsymbol{p}(\mathbf{c},\bar{t}(t,\mathbf{T})) - \hat{\boldsymbol{p}}\|_2^2$. 
According to Eqa.(\ref{equ: relative time})(\ref{equ: key point calculation}), for $i\in \left\{1,...,k-1 \right\}$, we can get the first order derivatives:
\begin{align}
	\frac{\partial f}{\partial \mathbf{c}_k}&=\frac{\partial f}{\partial \boldsymbol{p}}\frac{\partial \boldsymbol{p}}{\partial \mathbf{c}_k}=-2\boldsymbol{\beta}(\hat{\boldsymbol{p}}-\boldsymbol{p})^T,\\
	\frac{\partial f}{\partial T_i}&=\frac{\partial f}{\partial \boldsymbol{p}}\frac{\partial \boldsymbol{p}}{\partial \bar{t}}\frac{\partial \bar{t}}{\partial T_i}=-2(\hat{\boldsymbol{p}}-\boldsymbol{p})^T\dot{\boldsymbol{p}} \cdot(-1),\\	
	\frac{\partial f}{\partial t}&=\frac{\partial f}{\partial \boldsymbol{p}}\frac{\partial \boldsymbol{p}}{\partial t}+\frac{\partial f}{\partial \hat{\boldsymbol{p}}}\frac{\partial \hat{\boldsymbol{p}}}{\partial t}=2(\hat{\boldsymbol{p}}-\boldsymbol{p})^T(\dot{\hat{\boldsymbol{p}}}-\dot{\boldsymbol{p}}).
\end{align}
For the second order derivatives:
\begin{align}
	\frac{\partial^2 f}{\partial \mathbf{c}_k\partial t}&=-2\boldsymbol{\beta}(\dot{\hat{\boldsymbol{p}}}-\dot{\boldsymbol{p}})^T - 2\dot{\boldsymbol{\beta}}(\hat{\boldsymbol{p}}-\boldsymbol{p})^T,\\
	\frac{\partial^2 f}{\partial T_i\partial t}&=-2(-1)\cdot \Big((\hat{\boldsymbol{p}}-\boldsymbol{p})^T\ddot{\boldsymbol{p}}+(\dot{\hat{\boldsymbol{p}}}-\dot{\boldsymbol{p}})^T\boldsymbol{p}\Big), \\
	\frac{\partial^2f}{\partial t^2}&=2(\boldsymbol{p}-\hat{\boldsymbol{p}})^T(\ddot{\boldsymbol{p}}-\ddot{\hat{\boldsymbol{p}}})+2(\dot{\boldsymbol{p}}-\dot{\hat{\boldsymbol{p}}})^T(\dot{\boldsymbol{p}}-\dot{\hat{\boldsymbol{p}}}).
\end{align}
From (\ref{equ: bi-level gradients}), we can finally get the gradients of $t^*$ w.r.t $\mathbf{c},\mathbf{T}$:
\begin{align}
	\frac{\partial t^*}{\partial \mathbf{c}_k}&=-{\frac{\partial^2 f}{\partial \mathbf{c}_k\partial t}}/{\frac{\partial^2f}{\partial t^2}}, \\
	\frac{\partial t^*}{\partial T_i}&=-{\frac{\partial^2 f}{\partial T_i\partial t}}/{\frac{\partial^2f}{\partial t^2}}.
\end{align}

For the higher-level optimization problem, we should derive the gradients of $\mathcal{G}$ to $\mathbf{c},\mathbf{T}$.
As discussed above, $t^*$ is computed from (\ref{equ: inner optimization}), associated with $\mathbf{c},\mathbf{T}$.
Therefore, the layered representation of the penalty term $\mathcal{G}$ is:
\begin{align}
	\mathcal{G} = \mathcal{G}\Bigg(\boldsymbol{p}\bigg(\mathbf{c},\bar{t}\Big(t^*(\mathbf{c},\mathbf{T}),\mathbf{T}\Big)\bigg),\hat{\boldsymbol{p}},\dot{\boldsymbol{p}}\bigg(...\bigg),\dot{\hat{\boldsymbol{p}}} \Bigg),
\end{align}
where the variables in $\dot{\boldsymbol{p}}(...)$ are omitted since they are exactly the same as those in $\boldsymbol{p}$. 
Then we apply the chain rule:\looseness=-1
\begin{align}
	\frac{\partial \mathcal{G}}{\partial \mathbf{c}_k}&=\frac{\partial \mathcal{G}}{\partial \boldsymbol{p}}\frac{\partial \boldsymbol{p}}{\partial \mathbf{c}_k}+\frac{\partial \mathcal{G}}{\partial \dot{\boldsymbol{p}}}\frac{\partial \dot{\boldsymbol{p}}}{\partial \mathbf{c}_k}+\frac{\partial \mathcal{G}}{\partial t^*}\frac{\partial t^*}{\partial \mathbf{c}_k},\\
	\frac{\partial \mathcal{G}}{\partial T_i}&=\Big(\frac{\partial\mathcal{G}}{\partial \boldsymbol{p}}\frac{\partial \boldsymbol{p}}{\partial \bar{t}}+\frac{\partial\mathcal{G}}{\partial \dot{\boldsymbol{p}}}\frac{\partial \dot{\boldsymbol{p}}}{\partial \bar{t}}\Big)\frac{\partial \bar{t}}{\partial T_i} + \frac{\partial \mathcal{G}}{\partial t^*}\frac{\partial t^*}{\partial T_i}.
\end{align}
When the topological penalty term is greater than 0, according to Eqa.(\ref{equ: multi-vehicle homotopy invariant})(\ref{equ: topological penalty term}), we can get the higher-level gradients:
\begin{align}
	\frac{\partial \mathcal{G}}{\partial \boldsymbol{p}} &= \eta \mathbf{B}^T(\dot{\boldsymbol{p}} - \dot{\hat{\boldsymbol{p}}}), \\
	\frac{\partial \mathcal{G}}{\partial \dot{\boldsymbol{p}}} &= \eta \mathbf{B}(\boldsymbol{p} - \hat{\boldsymbol{p}}), \\
	\frac{\partial \mathcal{G}}{\partial t^*} &= \eta \big((\ddot{\boldsymbol{p}} - \ddot{\hat{\boldsymbol{p}}})^T \mathbf{B}(\boldsymbol{p} - \hat{\boldsymbol{p}}) + (\dot{\boldsymbol{p}} - \dot{\hat{\boldsymbol{p}}})^T \mathbf{B}(\dot{\boldsymbol{p}} - \hat{\dot{\boldsymbol{p}}}) \big).
\end{align}
By far we have gone through main derivations of both the higher-level and lower-level optimization in this bi-level trajectory optimization problem.

\subsection{Two-Stage Optimization}
The unconstrained optimization problem (\ref{equ: multi-vehicle trajectory optimization}) involves multiple penalty terms transformed from the constraints, including the topological and obstacle avoidance constraints.
In our previous work\cite{ma2023decentralized} and many other existing literature, the way to guarantee collision avoidance is to constrain the distance between vehicles longer than the safety threshold.
The penalty term is introduced to reflect the relationship between vehicle proximity and cost: the closer the vehicles are, the higher the cost becomes.
However, modifying the topological structure requires that the trajectory, while deforming from one side of the obstacle to the other, must first approach the obstacle and then move away from it, as shown in Fig. \ref{fig: topology deformation}.
This results in a problem that, when the trajectory approaches the obstacle, the penalty of collision avoidance increases whereas the penalty of topology decreases.
The conflict between collision avoidance and topological constraints is unavoidable during the optimization process, and forms a local minima where both constraints are not satisfied.

\begin{figure}[t]
	\centering
	\includegraphics[width=8.7cm]{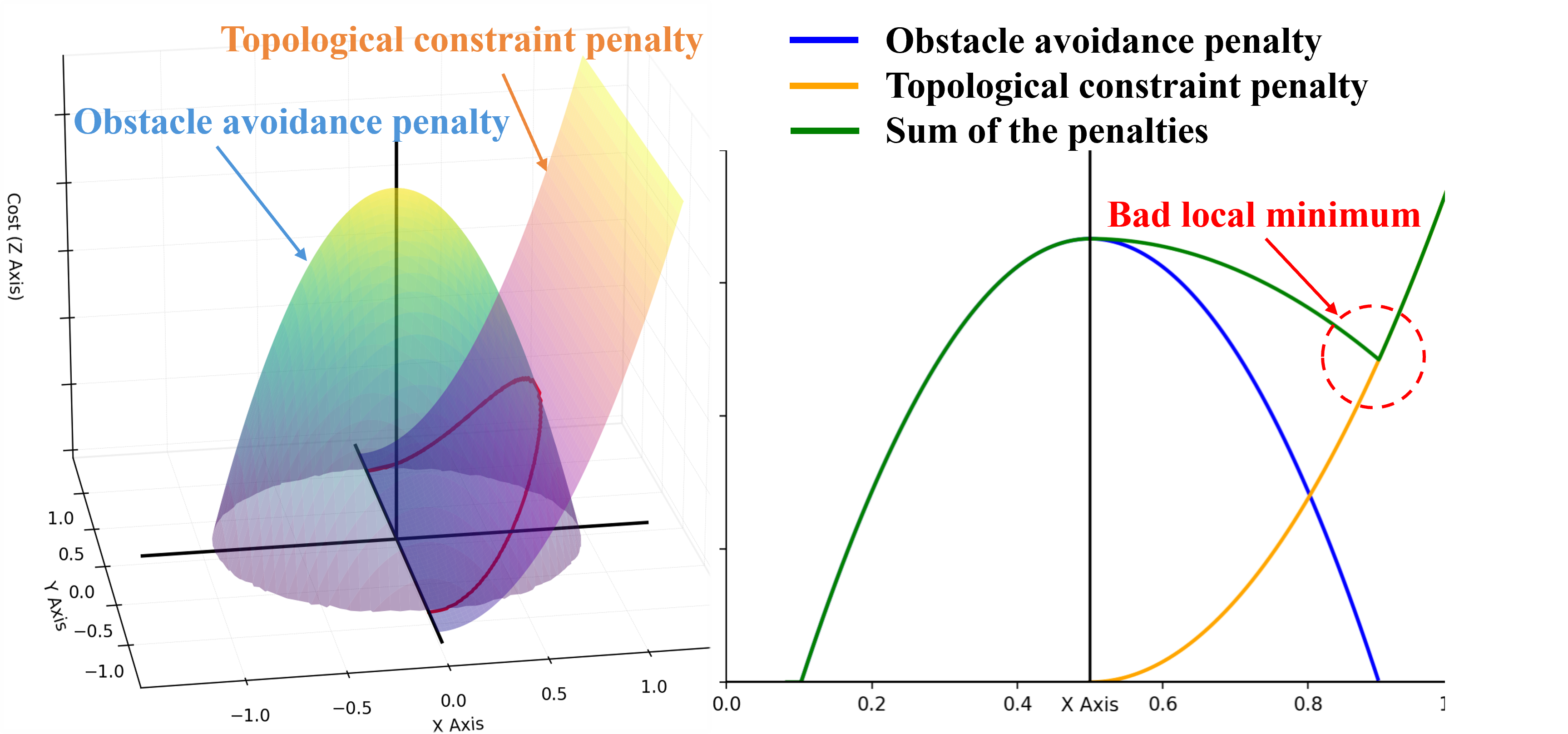}
	\setlength{\abovecaptionskip}{-0.4cm}
	\caption{Illustration of penalty terms. Two surfaces in the left figure respectively represent the obstacle avoidance penalty and topological constraint penalty. Right figure is a projection of the two surfaces onto the XOZ plane. The red dotted circle in the right figure is a bad local minimum caused by the conflict between obstacle avoidance and topological constraints.}
	\label{fig: penalty illustration}
	\vspace{-0.2cm}
\end{figure}

Fig. \ref{fig: penalty illustration} illustrates the penalty cost of the collision avoidance and topological constraints, where the local minima caused by their conflict is demonstrated.
To solve this problem, one may first think of setting the topology weight $w_t$ with a large value.
However, setting a large penalty weight might make the problem ill-conditioned, bring difficulty in fast convergence to the solution.
In this work, we divide the optimization problem into two stages.
In the first stage, we remove the collision avoidance penalty term and only retain the topological constraint term.
Once the topological constraint is satisfied, the first stage stopped and immediately runs into the second stage.
In the second stage, we add back the collision avoidance penalty term, and the optimization process proceeds until convergence.

Dividing the optimization into two stages, we avoid the conflict by only dealing with the topological constraint in the first stage, and focusing on the collision avoidance constraint in the second stage.
This strategy works well because in the second stage, the topological constraint has been satisfied with no penalty, and the gradients generated by collision avoidance penalty points to the direction where the topological constraint would not be violated.
Essentially, the first stage is similar to the trajectory sampling module in \cite{chen2023interactive,de2024topology}, whereas the difference is topology-guided optimization instead of random sampling.

\begin{figure*}[t]
	\centering
	\includegraphics[width=17.55cm]{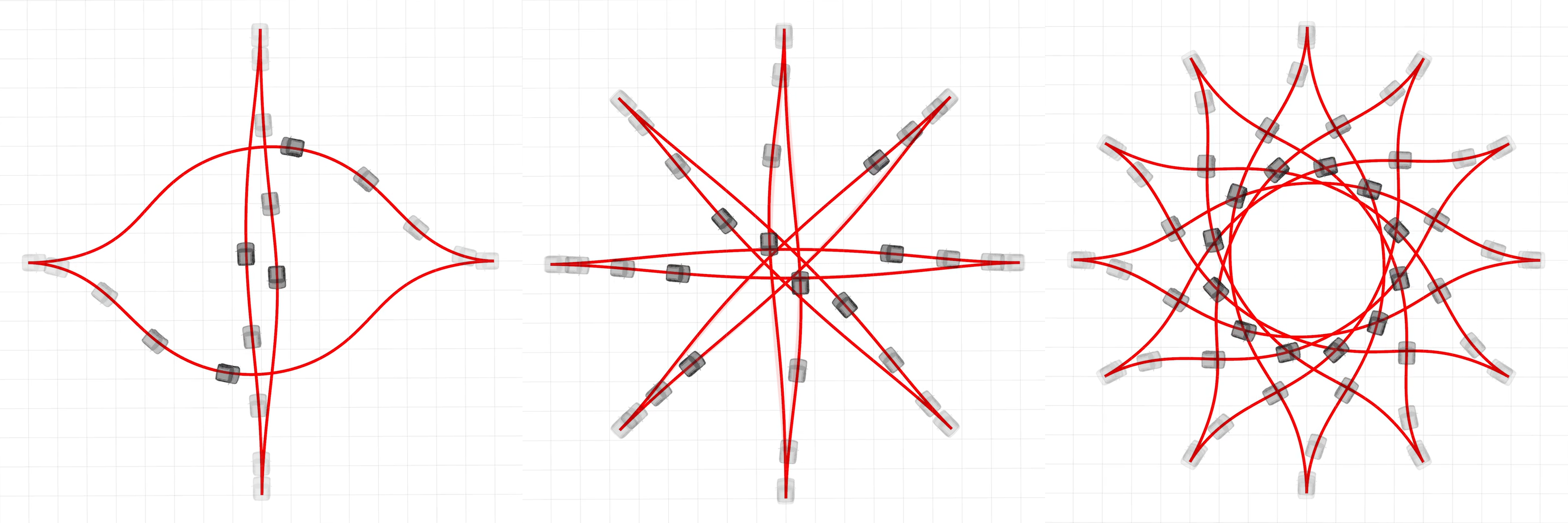}
	\setlength{\abovecaptionskip}{-0.0cm}
	\caption{Simulations about different interaction patterns within 4(left), 8(middle), 12(right) vehicles, respectively.}
	\label{fig: simulation experiments}
	\vspace{-0.0cm}
\end{figure*}

\begin{table*}
	\centering
	\centering
	\Large
	\renewcommand{\arraystretch}{1.2}
	\resizebox{1.0\textwidth}{!}{
		\begin{tabular}{l|l|ccc|ccc|ccc} 
			\toprule[2.5pt]
			\multicolumn{1}{l}{\multirow{2}{*}{}} & \multirow{2}{*}{Frameworks} & \multicolumn{3}{c|}{4 cars}                                                                                                                                                               & \multicolumn{3}{c|}{8 cars}                                                                                                                                                               & \multicolumn{3}{c}{12 cars}                                                                                                                                                                \\ 
			\cline{3-11}
			\multicolumn{1}{l}{}                  &                             & \begin{tabular}[c]{@{}c@{}}Computation\\time(ms)\end{tabular} & \begin{tabular}[c]{@{}c@{}}Travel\\Distance(m)\end{tabular} & \begin{tabular}[c]{@{}c@{}}Travel\\Duration(s)\end{tabular} & \begin{tabular}[c]{@{}c@{}}Computation\\Time(ms)\end{tabular} & \begin{tabular}[c]{@{}c@{}}Travel\\Distance(m)\end{tabular} & \begin{tabular}[c]{@{}c@{}}Travel\\Duration(s)\end{tabular} & \begin{tabular}[c]{@{}c@{}}Computation\\Time(ms)\end{tabular} & \begin{tabular}[c]{@{}c@{}}Travel\\Distance(m)\end{tabular} & \begin{tabular}[c]{@{}c@{}}Travel\\Duration(s)\end{tabular}  \\ 
			\hline
			\multirow{4}{*}{\rotcell{set 1}} & Direct Opt\cite{10325486}                  & \textbf{310}                                                  & 64.43                                                       & 33.5                                                        & 1023                                                          & 129.21                                                      & 80.9                                                        & \textbf{2408}                                                 & 193.55                                                      & 140.8                                                        \\
			& Priority-based\cite{Li_2021}              & 457                                                           & 64.41                                                       & 33.1                                                        & 1727                                                          & 130.8                                                       & 81.6                                                        & 4056                                                          & 199.58                                                      & 141.6                                                        \\
			& Winding Angle\cite{chen2023interactive}               & 2985                                                          & 64.44                                                       & 40.0                                                        & 162236                                                        & 132.44                                                      & 100                                                         & /                                                             & /                                                           & /                                                            \\
			& \textbf{Proposed}           & 351                                                           & \textbf{64.38}                                              & \textbf{32.3}                                               & \textbf{959}                                                  & \textbf{128.84}                                             & \textbf{78.2}                                               & 2582                                                          & \textbf{193.37}                                             & \textbf{136.5}                                               \\ 
			\hline\hline
			\multirow{3}{*}{\rotcell{set 2}}      & Priority-based\cite{Li_2021}              & 354                                                           & 65.43                                                       & \textbf{29.8}                                               & 1054                                                          & 134.07                                                      & 74.6                                                        & 5243                                                          & 215.34                                                      & 124.1                                                        \\
			& Winding Angle\cite{chen2023interactive}               & 3277                                                          & 64.91                                                       & 40.0                                                        & 8679                                                          & 134.65                                                      & 100.0                                                       & /                                                             & /                                                           & /                                                            \\
			& \textbf{Proposed}           & \textbf{163}                                                  & \textbf{64.75}                                              & 30.3                                                        & \textbf{684}                                                  & \textbf{132.23}                                             & \textbf{68.9}                                               & \textbf{2085}                                                 & \textbf{205.27}                                             & \textbf{106.3}                                               \\
			\bottomrule
		\end{tabular}
	}
	\caption{Benchmark comparisons with existing trajectory optimization frameworks under multiple interactions.}
	\label{tab: benchmark comparisons}
	\vspace{-0.6cm}
\end{table*}

\begin{figure}[t]
	\centering
	\includegraphics[width=8.5cm]{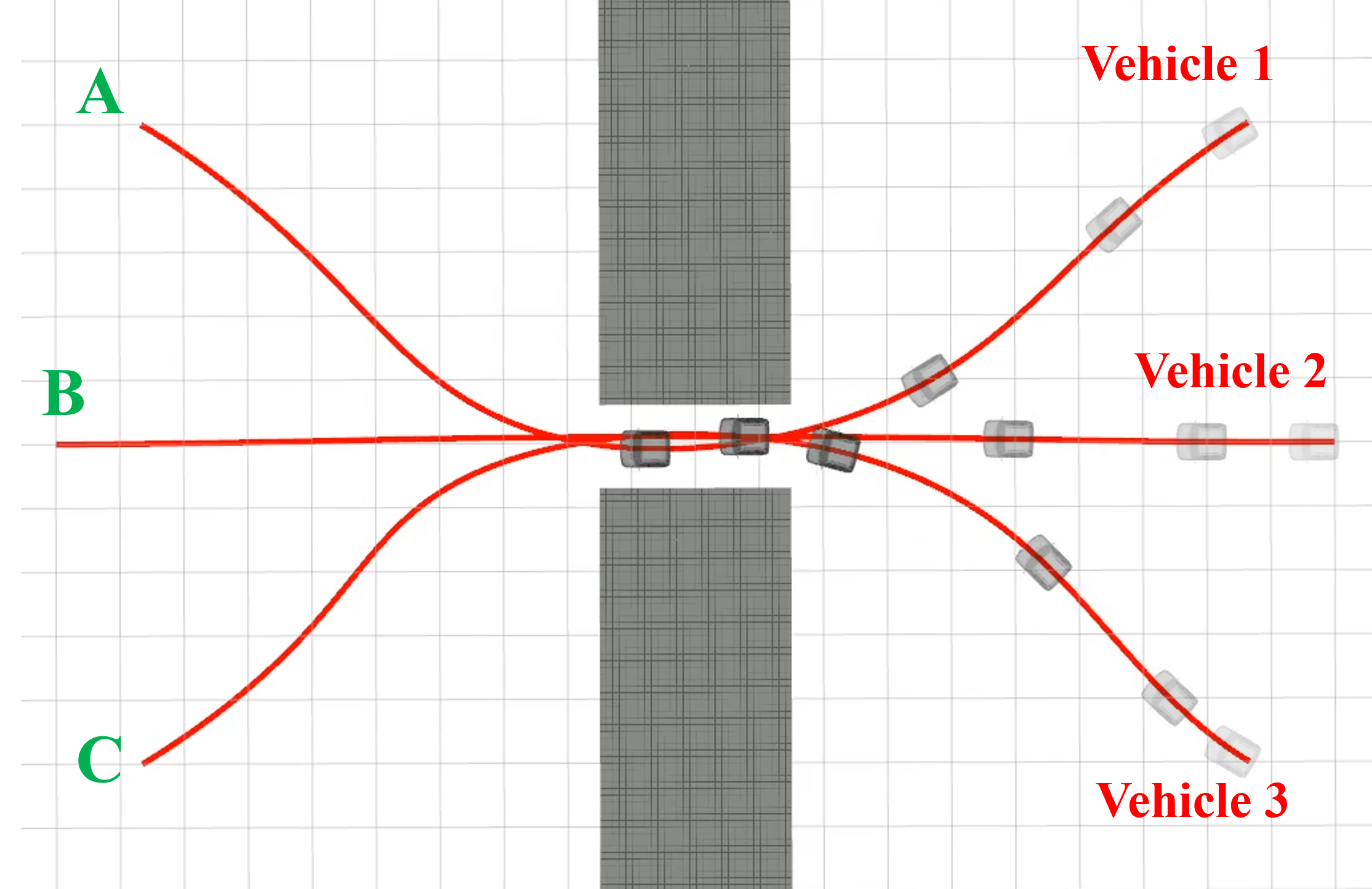}
	\setlength{\abovecaptionskip}{-0.1cm}
	\caption{Illustration of three vehicles traversing through the narrow corridor one by one. The goals of vehicles 1, 2, 3 are points A, B, C, respectively. In this scenario, we set the vehicles move counterclockwise relative to each other. Therefore, vehicle 1, 2, 3 sequentially passes through the narrow corridor under this interaction pattern.}
	\label{fig: narrow corridor}
	\vspace{-0.3cm}
\end{figure}

\section{Results and Experiments}
In this section, we first introduce some implementation details.
Then we conduct some simulations and benchmark comaprisons.
Real-world experiments are also conducted to further validate the effectiveness.

\subsection{Implementation Details}
Some implementation details are mentioned here.
The calculation of the key point timestamp (\ref{equ: inner optimization}) is based on a simple gradient descent process.
The overall optimization problem (\ref{equ: multi-vehicle trajectory optimization}) is solved by the L-BFGS algorithm\cite{liu1989limited}.
The project is implemented using C++ in a ROS environment, equipped with an Intel i9-13900K CPU.

Besides, the penalty weight $w_T$ on total duration is set as 100.
The penalty weight $w_t$ on topological constraint term differs in two stages: 500 in the first stage and 5000 in the second stage.

\subsection{Simulations and Benchmark Comparisons}
We conduct massive simulation and benchmark comparison experiments to validate the superiority of the proposed algorithm.
We compare our project with existing state-of-the-art multi-vehicle trajectory optimization framework\cite{10325486}, priority-based implicit interactive planning framework\cite{Li_2021}, and topological planning system based on the winding angle\cite{chen2023interactive}.
The comparison metrics we care about include time consumption in computation, the overall travel length and time duration.
We test these with 4, 8, 12 vehicles under different types of interactions, respectively shown in Fig. \ref{fig: simulation experiments}.
All simulations are conducted in a $10m$ $\times$ $10m$ environment.
The vehicles obey Ackerman car models in size of $0.85m$ $\times$ $0.65m$.
The speed and acceleration limits are set as $3m/s$ and $2m/s^2$.
To ensure fairness, all frameworks are provided with the same initial values before optimization. 

The results are listed in Tab.\ref{tab: benchmark comparisons}.
The table is divided by a double-line in the row direction, which means we conducted two sets of experiments.
The first set corresponds to the data above the double-line in the table.
Since framework \cite{10325486} relies on direct optimization without the ability to control the interactions, we first run this framework and memorize the interaction relationship between vehicles.
And then we fix this interaction pattern and respectively run priority-based trajectory planning framework\cite{Li_2021}, winding angle-based interactive optimization method\cite{chen2023interactive}, as well as the proposed framework.
The second set corresponds to the data below the double-line in the table.
We eliminate the direct optimization method, and design some other kinds of interactions.
Then we execute the trajectory optimization under these interactions using the rest of the frameworks.
We count on the average comparison metrics and record them in this table.
Parts of interaction patterns are shown in Fig. \ref{fig: simulation experiments}.
Note that data of \cite{chen2023interactive} is not available when there are 12 vehicles.
This is because when the number of the vehicles is growing, the Forces Pro\cite{FORCESNLP} client fails to generate the solver under heavy computational burden, hindering further test with more vehicles.

By analyzing the comparison results, we can find the proposed framework reaches better performance in both optimality and efficiency than existing frameworks.
In the first set of experiments, all frameworks are running in the same interaction pattern as \cite{10325486}.
Compared to the state-of-the-art\cite{10325486} on multi-vehicle trajectory optimization, the proposed framework shows a slight disadvantage in computational efficiency but demonstrates superior optimality.
This is due to the bi-level optimization process embedded in our iterative computations, which introduces additional computational overhead.
Nevertheless, this slight trade-off in efficiency is well justified by the enhanced flexibility and controllability over the interactions that our framework enables.
In the second set of experiments, we modify several interaction patterns.
Under interaction relationships that are different from those in the first set of experiments, the proposed framework still outperforms the existing algorithms.
It's worth mentioning that the optimization of \cite{chen2023interactive} fails under some specific interactions, where the solver returns failure due to infinite values.
This phenomenon probably results from the Non-differentiability of the winding angle, which results in infinite gradients during the optimization process.

In addition, we conduct experiments in an environment with static obstacles, as shown in Fig. \ref{fig: narrow corridor}.
In this figure, three vehicles manage to sequentially pass through the narrow corridor by constraining the specific topological relationships.
Suppose the vehicle 1 runs towards point A, vehicle 2 towards point B, vehicle 3 towards point C.
In this scenario, no matter which vehicle passes through corridor first, the final winding angles between trajectories are not varying.
This is because the trajectories in this scenario are not necessarily crossing each other, and the winding angle focuses on the complete process from start to goal.
Therefore, the winding angle is not adequate to describe the interactions in this scenario.
On the contrary, the proposed topological metric only focuses on the local features at the nearest points.
This property gives it chances to analyze the interactions at the narrow corridor, making controllable interactions possible.

Overall, the benchmark comparisons in the simulation experiments demonstrate both the optimality and computation efficiency over existing frameworks.

\subsection{Real-world Experiments}
We also conduct real-world experiments in a $10m \times 10m$ environment, where four Ackerman vehicles move from one side to the other.
Each vehicle is equipped with an NVIDIA Jetson Orin as the onboard processor for computation and a lidar for localization.
Since the proposed algorithm is a centralized global planning framework, we deploy a central computer for trajectory optimization for all vehicles.
Then the computation results are transmitted to the onboard processors over Wi-Fi module.
After that, the vehicles execute the received trajectories through the controllers.

Maximum speed and acceleration are respectively set as $1.5m/s$ and $0.8m/s^2$.
Fig. \ref{fig:real world experiments} illustrates the real-world experiments where four vehicles navigates counterclockwise at the interaction area.
We also conduct real-world experiments under different interaction patterns.
For more details, please watch the attached video.

\section{Conclusion and Discussion}
In this work, we focus on achieving controllable interactions in multi-vehicle scenarios, addressing the limitations of existing methods that lack explicit controllability over vehicle interactions during trajectory optimization. We propose a differentiable homotopy invariant metric to model interactions, which is integrated as a constraint into the optimization framework. This enables the generation of multiple interaction patterns from the same initial values, providing users with the flexibility to design and control vehicle interactions. Extensive experiments demonstrate the effectiveness of our approach in achieving controllable and diverse interactions while maintaining computational efficiency. By releasing open-source code, we aim to support further research in multi-vehicle interactive planning.

While our framework demonstrates significant advancements in modeling controllable interactions, it is not without limitations. The local homotopy invariant metric focuses on interactions at the nearest points, which may not fully capture more complex scenarios such as multiple sequential interactions or looping trajectories. These cases present unique challenges that remain to be addressed. Moving forward, we aim to refine our approach to better handle such complexities and expand the applicability of our framework to a broader spectrum of real-world scenarios.


\begin{thebibliography}{10}
	\providecommand{\url}[1]{#1}
	\csname url@rmstyle\endcsname
	\providecommand{\newblock}{\relax}
	\providecommand{\bibinfo}[2]{#2}
	\providecommand\BIBentrySTDinterwordspacing{\spaceskip=0pt\relax}
	\providecommand\BIBentryALTinterwordstretchfactor{4}
	\providecommand\BIBentryALTinterwordspacing{\spaceskip=\fontdimen2\font plus
		\BIBentryALTinterwordstretchfactor\fontdimen3\font minus \fontdimen4\font\relax}
	\providecommand\BIBforeignlanguage[2]{{%
			\expandafter\ifx\csname l@#1\endcsname\relax
			\typeout{** WARNING: IEEEtran.bst: No hyphenation pattern has been}%
			\typeout{** loaded for the language `#1'. Using the pattern for}%
			\typeout{** the default language instead.}%
			\else
			\language=\csname l@#1\endcsname
			\fi
			#2}}
	
	\bibitem{bhattacharya2010search}
	S.~Bhattacharya, ``Search-based path planning with homotopy class constraints,'' in \emph{Proceedings of the AAAI conference on artificial intelligence}, vol.~24, no.~1, 2010, pp. 1230--1237.
	
	\bibitem{7182335}
	J.~Park, S.~Karumanchi, and K.~Iagnemma, ``Homotopy-based divide-and-conquer strategy for optimal trajectory planning via mixed-integer programming,'' \emph{IEEE Transactions on Robotics}, vol.~31, no.~5, pp. 1101--1115, 2015.
	
	\bibitem{sontges2017computing}
	S.~S{\"o}ntges and M.~Althoff, ``Computing possible driving corridors for automated vehicles,'' in \emph{2017 IEEE Intelligent Vehicles Symposium (IV)}.\hskip 1em plus 0.5em minus 0.4em\relax IEEE, 2017, pp. 160--166.
	
	\bibitem{bhattacharya2018path}
	S.~Bhattacharya and R.~Ghrist, ``Path homotopy invariants and their application to optimal trajectory planning,'' \emph{Annals of Mathematics and Artificial Intelligence}, vol.~84, no.~3, pp. 139--160, 2018.
	
	\bibitem{6425970}
	S.~Kim, K.~Sreenath, S.~Bhattacharya, and V.~Kumar, ``Optimal trajectory generation under homology class constraints,'' in \emph{2012 IEEE 51st IEEE Conference on Decision and Control (CDC)}, 2012, pp. 3157--3164.
	
	\bibitem{mavrogiannis2022analyzing}
	C.~Mavrogiannis, J.~DeCastro, and S.~S. Srinivasa, ``Analyzing multiagent interactions in traffic scenes via topological braids,'' in \emph{2022 International Conference on Robotics and Automation (ICRA)}.\hskip 1em plus 0.5em minus 0.4em\relax IEEE, 2022, pp. 5806--5813.
	
	\bibitem{liu2024betop}
	H.~Liu, L.~Chen, Y.~Qiao, C.~Lv, and H.~Li, ``Reasoning multi-agent behavioral topology for interactive autonomous driving,'' in \emph{NeurIPS}, 2024.
	
	\bibitem{doi:10.1177/02783649231188740}
	\BIBentryALTinterwordspacing
	C.~Mavrogiannis, J.~A. DeCastro, and S.~S. Srinivasa, ``Abstracting road traffic via topological braids: Applications to traffic flow analysis and distributed control,'' \emph{The International Journal of Robotics Research}, vol.~43, no.~9, pp. 1299--1321, 2024. [Online]. Available: \url{https://doi.org/10.1177/02783649231188740}
	\BIBentrySTDinterwordspacing
	
	\bibitem{jaillet2008path}
	L.~Jaillet and T.~Sim{\'e}on, ``Path deformation roadmaps: Compact graphs with useful cycles for motion planning,'' \emph{The International Journal of Robotics Research}, vol.~27, no. 11-12, pp. 1175--1188, 2008.
	
	\bibitem{zhou2021raptor}
	B.~Zhou, J.~Pan, F.~Gao, and S.~Shen, ``Raptor: Robust and perception-aware trajectory replanning for quadrotor fast flight,'' \emph{IEEE Transactions on Robotics}, vol.~37, no.~6, pp. 1992--2009, 2021.
	
	\bibitem{zhou2022swarm}
	X.~Zhou, X.~Wen, Z.~Wang, Y.~Gao, H.~Li, Q.~Wang, T.~Yang, H.~Lu, Y.~Cao, C.~Xu, \emph{et~al.}, ``Swarm of micro flying robots in the wild,'' \emph{Science Robotics}, vol.~7, no.~66, p. eabm5954, 2022.
	
	\bibitem{bhattacharya2012topological}
	S.~Bhattacharya, M.~Likhachev, and V.~Kumar, ``Topological constraints in search-based robot path planning,'' \emph{Autonomous Robots}, vol.~33, pp. 273--290, 2012.
	
	\bibitem{gu2016automated}
	T.~Gu, J.~M. Dolan, and J.-W. Lee, ``Automated tactical maneuver discovery, reasoning and trajectory planning for autonomous driving,'' in \emph{2016 IEEE/RSJ international conference on intelligent robots and systems (IROS)}.\hskip 1em plus 0.5em minus 0.4em\relax IEEE, 2016, pp. 5474--5480.
	
	\bibitem{baselga2024shine}
	D.~M. Baselga, O.~de~Groot, L.~Kn{\"o}dler, L.~Riazuelo, J.~Alonso-Mora, and L.~Montano, ``Shine: Social homology identification for navigation in crowded environments,'' \emph{CoRR}, 2024.
	
	\bibitem{inbook}
	P.~Vernaza, V.~Narayanan, and M.~Likhachev, \emph{Efficiently Finding Optimal Winding-Constrained Loops in the Plane}, 07 2013, pp. 417--424.
	
	\bibitem{kretzschmar2016socially}
	H.~Kretzschmar, M.~Spies, C.~Sprunk, and W.~Burgard, ``Socially compliant mobile robot navigation via inverse reinforcement learning,'' \emph{The International Journal of Robotics Research}, vol.~35, no.~11, pp. 1289--1307, 2016.
	
	\bibitem{roh2021multimodal}
	J.~Roh, C.~Mavrogiannis, R.~Madan, D.~Fox, and S.~Srinivasa, ``Multimodal trajectory prediction via topological invariance for navigation at uncontrolled intersections,'' in \emph{Conference on Robot Learning}.\hskip 1em plus 0.5em minus 0.4em\relax PMLR, 2021, pp. 2216--2227.
	
	\bibitem{mavrogiannis2021hamiltonian}
	C.~Mavrogiannis and R.~A. Knepper, ``Hamiltonian coordination primitives for decentralized multiagent navigation,'' \emph{The International Journal of Robotics Research}, vol.~40, no. 10-11, pp. 1234--1254, 2021.
	
	\bibitem{mavrogiannis2022winding}
	C.~Mavrogiannis, K.~Balasubramanian, S.~Poddar, A.~Gandra, and S.~S. Srinivasa, ``Winding through: Crowd navigation via topological invariance,'' \emph{IEEE Robotics and Automation Letters}, vol.~8, no.~1, pp. 121--128, 2022.
	
	\bibitem{chen2023interactive}
	Y.~Chen, S.~Veer, P.~Karkus, and M.~Pavone, ``Interactive joint planning for autonomous vehicles,'' \emph{IEEE Robotics and Automation Letters}, 2023.
	
	\bibitem{ma2023decentralized}
	C.~Ma, Z.~Han, T.~Zhang, J.~Wang, L.~Xu, C.~Li, C.~Xu, and F.~Gao, ``Decentralized planning for car-like robotic swarm in cluttered environments,'' in \emph{2023 IEEE/RSJ International Conference on Intelligent Robots and Systems (IROS)}.\hskip 1em plus 0.5em minus 0.4em\relax IEEE, 2023, pp. 9293--9300.
	
	\bibitem{10285583}
	Z.~Han, Y.~Wu, T.~Li, L.~Zhang, L.~Pei, L.~Xu, C.~Li, C.~Ma, C.~Xu, S.~Shen, and F.~Gao, ``An efficient spatial-temporal trajectory planner for autonomous vehicles in unstructured environments,'' \emph{IEEE Transactions on Intelligent Transportation Systems}, vol.~25, no.~2, pp. 1797--1814, 2024.
	
	\bibitem{10325486}
	L.~Pei, J.~Lin, Z.~Han, L.~Quan, Y.~Cao, C.~Xu, and F.~Gao, ``Collaborative planning for catching and transporting objects in unstructured environments,'' \emph{IEEE Robotics and Automation Letters}, vol.~9, no.~2, pp. 1098--1105, 2024.
	
	\bibitem{de2024topology}
	O.~De~Groot, L.~Ferranti, D.~M. Gavrila, and J.~Alonso-Mora, ``Topology-driven parallel trajectory optimization in dynamic environments,'' \emph{IEEE Transactions on Robotics}, 2024.
	
	\bibitem{Li_2021}
	\BIBentryALTinterwordspacing
	J.~Li, M.~Ran, and L.~Xie, ``Efficient trajectory planning for multiple non-holonomic mobile robots via prioritized trajectory optimization,'' \emph{IEEE Robotics and Automation Letters}, vol.~6, no.~2, p. 405–412, Apr. 2021. [Online]. Available: \url{http://dx.doi.org/10.1109/LRA.2020.3044834}
	\BIBentrySTDinterwordspacing
	
	\bibitem{huang2024spatiotemporal}
	Y.~Huang, W.~He, Y.~Kantaros, and S.~Zeng, ``Spatiotemporal co-design enabling prioritized multi-agent motion planning,'' in \emph{2024 IEEE/RSJ International Conference on Intelligent Robots and Systems (IROS)}.\hskip 1em plus 0.5em minus 0.4em\relax IEEE, 2024, pp. 10\,281--10\,288.
	
	\bibitem{10533430}
	X.~Zhang, G.~Xiong, Y.~Wang, S.~Teng, and L.~Chen, ``D-pbs: Dueling priority-based search for multiple nonholonomic robots motion planning in congested environments,'' \emph{IEEE Robotics and Automation Letters}, vol.~9, no.~7, pp. 6288--6295, 2024.
	
	\bibitem{wang2022geometrically}
	Z.~Wang, X.~Zhou, C.~Xu, and F.~Gao, ``Geometrically constrained trajectory optimization for multicopters,'' \emph{IEEE Transactions on Robotics}, 2022.
	
	\bibitem{gould2016differentiating}
	S.~Gould, B.~Fernando, A.~Cherian, P.~Anderson, R.~S. Cruz, and E.~Guo, ``On differentiating parameterized argmin and argmax problems with application to bi-level optimization,'' \emph{arXiv preprint arXiv:1607.05447}, 2016.
	
	\bibitem{liu1989limited}
	D.~C. Liu and J.~Nocedal, ``On the limited memory bfgs method for large scale optimization,'' \emph{Mathematical programming}, vol.~45, no.~1, pp. 503--528, 1989.
	
	\bibitem{FORCESNLP}
	A.~Zanelli, A.~Domahidi, J.~Jerez, and M.~Morari, ``Forces nlp: an efficient implementation of interior-point... methods for multistage nonlinear nonconvex programs,'' \emph{International Journal of Control}, pp. 1--17, 2017.
	
\end{thebibliography}

\end{document}